\documentclass[10 pt, journal]{IEEEtran}

\usepackage{amsmath, amssymb, amsthm}
\usepackage{hyperref, graphicx, cite, multirow}
\usepackage[caption = false, font = footnotesize]{subfig}
\usepackage{pgfplots, pgfplotstable}
\usepgfplotslibrary{external, statistics}
\usepackage{tikz}
\usetikzlibrary{shapes, arrows, positioning, plotmarks, patterns}

\definecolor{turquoise}{RGB}{0, 120, 200}

\hypersetup{
colorlinks = true,
linkcolor = turquoise,
citecolor = turquoise,
filecolor = turquoise,
urlcolor = turquoise,
bookmarks = true,
pdftoolbar = true,
pdfmenubar = true,
pdfstartview = {FitH},
pdftitle = {Estimating the Probability that a Vehicle Reaches a Near-Term Goal State Using Multiple Lane Changes},
pdfauthor = {Goodarz Mehr},
pdfsubject = {Lane-Change Probability Model},
}

\newtheorem{theorem}{Theorem}
\newtheorem{lemma}{Lemma}

\makeatletter
\let\@ORGmakecaption\@makecaption
\long\def\@makecaption#1#2{\@ORGmakecaption{#1}{#2}\vskip\belowcaptionskip\relax}
\makeatother

\begin{document}

	\title{Estimating the Probability that a Vehicle Reaches a Near-Term Goal State Using Multiple Lane Changes}
	
	\author{Goodarz Mehr, and Azim Eskandarian, \IEEEmembership{Senior Member, IEEE}\thanks{Manuscript received February 19, 2020; revised October 8, 2020; accepted January 13, 2021. \\The authors are with the Autonomous Systems and Intelligent Machines (ASIM) Lab, Virginia Tech, Blacksburg, VA 24060, USA (email: \href{mailto:goodarzm@vt.edu}{goodarzm@vt.edu}; \href{mailto:eskandarian@vt.edu}{eskandarian@vt.edu}).\\\copyright 2021 IEEE. Personal use of this material is permitted.  Permission from IEEE must be obtained for all other uses, in any current or future media, including reprinting/republishing this material for advertising or promotional purposes, creating new collective works, for resale or redistribution to servers or lists, or reuse of any copyrighted component of this work in other works.}}
	
	\markboth{IEEE Transactions on Intelligent Transportation Systems}{Mehr \MakeLowercase{\textit{et. al.}}: Estimating the Probability that a Vehicle Reaches a Near-Term Goal State Using Multiple Lane Changes}
	
	\IEEEpubid{0000--0000/00\$00.00 \copyright\,2021 IEEE}
	
	\IEEEaftertitletext{\vspace{-4 pt}}
	
	\maketitle
	
	
	\begin{abstract}
		
		This paper proposes a model to estimate the probability of a vehicle reaching a near-term goal state using one or multiple lane changes based on parameters corresponding to traffic conditions and driving behavior. The proposed model not only has broad application in path planning and autonomous vehicle navigation, it can also be incorporated in advance warning systems to reduce traffic delay during recurrent and non-recurrent congestion. The model is first formulated for a two-lane road segment through systemic reduction of the number of parameters and transforming the problem into an abstract statistical form, for which the probability can be calculated numerically. It is then extended to cases with a higher number of lanes using the law of total probability. VISSIM\textsuperscript{\textregistered} simulations are used to validate the predictions of the model and study the effect of different parameters on the probability. For most cases, simulation results are within 4\% of model predictions, and the effect of different parameters such as driving behavior and traffic density on the probability match our expectation. The model can be implemented with near real-time performance, with computation time increasing linearly with the number of lanes.
	
	\end{abstract}
	
	\section*{Nomenclature}
	\addcontentsline{toc}{section}{Nomenclature}
	
	\begin{IEEEdescription}[\IEEEusemathlabelsep\IEEEsetlabelwidth{$P(S)$}]
		\item[$d$] Distance to the goal state.
		\item[$d_{e}$] Distance to search for the critical gap.
		\item[$d_{i}$] Maximum possible distance for initiating a successful lane changing maneuver.
		\item[$d_{r}$] Distance traveled by the ego vehicle relative to its adjacent lane.
		\item[erf] Gaussian error function.
		\item[$f, q$] Probability function.
		\item[$F(h)$] Cumulative distribution function of headway distance random variables.
		\item[$g$] Minimum acceptable (critical) gap.
		\item[$H_{i, k}$] Random variable representing $k$-th headway distance on lane $i$.
		\item[$i, j, k$] Lane index.\\Constant.
		\item[$l$] Index of adjacent lane leading vehicle.
		\item[$m$] Index of mean variable (velocity).
		\item[$n$] Number of lanes.
		\item[$P(S)$] Estimated success probability of reaching the goal state.
		\item[$P_{t}$] True success probability of reaching the goal state.
		\item[$s_{0}$] Minimum standstill distance.
		\item[$t$] Duration of a lane changing maneuver.\\Index of adjacent lane trailing vehicle.
		\item[$v$] Average vehicle velocity.
		\item[$V$] Desired velocity.
		\item[$x$] Location of completing a lane changing maneuver.
		\item[$X, Y$] Log-normal random variables.
		\item[$\delta$] Minimum desired headway in adjacent lane.
		\item[$\mu$] Natural logarithm mean of a log-normal random variable.
		\item[$\rho$] Traffic density (flow).
		\item[$\rho_{l}$] Traffic density per lane.
		\item[$\sigma$] Natural logarithm standard deviation of a log-normal random variable.
	\end{IEEEdescription}
	
	\section{Introduction} \label{Introduction}
	
	\IEEEPARstart{L}{ane} changes are an essential part of driving. Each maneuver depends on a multitude of factors such as the purpose and urgency of changing lanes, state of nearby vehicles, and driving behavior \cite{Brackstone}. A successful maneuver requires the driver (or autonomous vehicle) to identify a suitable gap in the target lane, adjust speed and maintain correct position relative to nearby vehicles, and navigate to the target lane while avoiding collision with other vehicles \cite{Kesting}. A small mistake at any step of this maneuver or unsafe driving behavior can result in an accident. In the United States, unsafe lane changing behavior is the cause of around four to ten percent of all reported motor vehicle crashes. In addition to the fatalities, such incidents incur an economic loss by creating congestion that delays traffic \cite{Sen, Lisheng, Dijck}. By one estimate, congestion cost the U.S. an estimated \$166 billion in 2019 \cite{UrbanMob}, with traffic incidents responsible for about a quarter of total delay on U.S. roadways \cite{NTIMC}. Such incidents and delays can be potentially mitigated if vehicles obtain accurate and timely information to avoid rushed lane changes.
	
	\IEEEpubidadjcol
	
	Lane changes are classified as either discretionary or mandatory \cite{Zhang1}. Discretionary lane changes are often performed to overtake slow traffic and move to a lane with a higher speed. In contrast, mandatory lane changes are required to achieve a navigation objective, for example to reach a highway off-ramp. Compared to discretionary lane changes, mandatory lane changes can have a disruptive impact on traffic. For example, mandatory lane changes during congestion can cause capacity drop \cite{Cassidy}, traffic oscillation \cite{Sarvi}, traffic breakdown \cite{Lv}, and deteriorate traffic safety \cite{Ahammed, Li}.
	
	\begin{figure*}[t!]
		\centering
		\includegraphics[width = \textwidth]{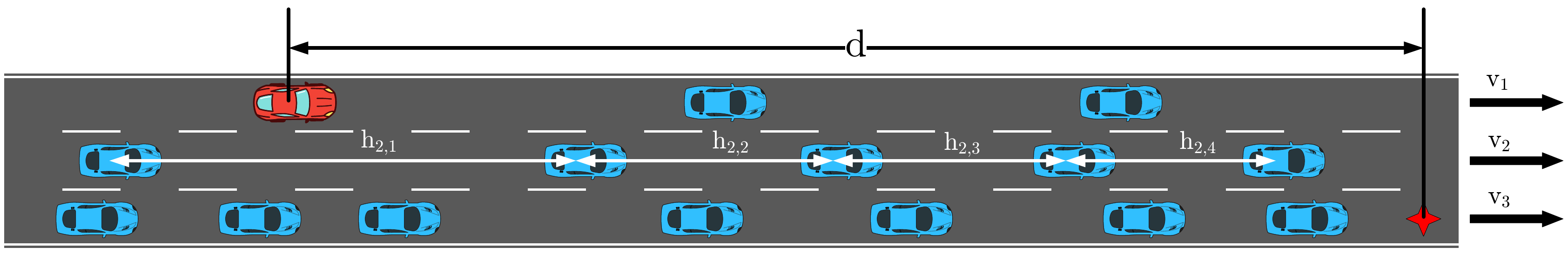}
		\setlength{\abovecaptionskip}{-11 pt}
		\caption{Notations used throughout this paper for a road segment with three lanes. The red car is the ego vehicle and the red star indicates the goal state \cite{Mehr1, Mehr3}.} \label{Nomenclature}
	\end{figure*}
	\setlength{\abovecaptionskip}{0.5\baselineskip}
	
	Several studies over the past decade have developed algorithms that help autonomous vehicles perform safe lane changes \cite{Atagoziyev, Cesari, Chandra, Iberraken, Liu1, Liu2, Vallon, Wang}. The focus of these studies has been on developing rule-based or machine learning-based methods and control algorithms that determine the trajectory to follow during the maneuver and help avoid collisions. For example, \cite{Chandra} proposed a method to alter the critical safety gap in the target lane by taking advantage of braking and steering actions without compromising safety, while \cite{Vallon} proposed integrating a support vector machine (SVM)-based method to initiate a lane change into a model predictive control (MPC) framework to create a more robust and personalized lane changing experience. Although these methods can handle a safe lane change from beginning to end, they are generally unable to determine the right time to initiate a lane change to reach a target position on another (perhaps non-adjacent) lane that minimizes traffic disruption. Take, for example, an autonomous vehicle traveling in the leftmost lane of a four-lane highway that needs to take an off-ramp. Current methods cannot determine a good time to start changing lanes to make sure that by the time the vehicle reaches the off-ramp, it is in the rightmost lane. If the lane change is initiated too early, the vehicle has to spend considerable time traveling at a potentially lower speed in the rightmost lane. Conversely, if the maneuver is initiated too late, the vehicle may miss the exit entirely, or may be forced to do multiple rushed lane changes, potentially slowing down nearby traffic and increasing the likelihood of an incident \cite{Gong}.
	
	Based on studies conducted on human drivers, advance warning systems are a potential way to address this problem for autonomous vehicles and decrease the number of rushed and unsafe lane changes \cite{Gong, Hang, He}. \cite{Hang} used a driving simulator to study the effects of advance warning location in work zone areas on lane changing behavior and found that it had a strong impact on drivers' perception of the imminent situation. Similarly, \cite{He} found that providing timely advance warning can reduce average travel time, especially in moderate and congested traffic. At the core of this approach is developing a model that can tell an autonomous vehicle the likelihood of successfully performing one or multiple lane changes to reach a target position ahead. While some studies have developed such probability models, they all have some drawbacks \cite{Gong, He, Luo}. Some models only work with a particular type of traffic (for example exponential headway distribution) while others cannot be deployed in real time. Most importantly, all derivations are limited to only two lanes, which is not practical in many situations and especially for highway driving. \par
	
	This paper addresses the limitations of previous studies. Specifically, we formulate a model that can accurately estimate the probability of reaching a certain goal state in a different lane using one or multiple lane changes under general traffic conditions \cite{Mehr1}. Model accuracy is validated through VISSIM\textsuperscript{\textregistered} simulations and real-time performance is demonstrated through MATLAB\textsuperscript{\textregistered} profiling. As discussed above, the proposed model not only has broad application in path planning and autonomous vehicle navigation, as shown in \cite{Mehr2, Mehr3, Mehr4} it can also be incorporated in advance warning systems to reduce traffic delay during recurrent and non-recurrent congestion \cite{VanDriel, Roncoli, Zhang2, Zhang3}. Broad adoption of such systems can potentially save billions in costs associated with congestion. \par
	
	The remainder of this paper is structured as follows. \autoref{Problem} presents problem formulation and assumptions made to develop the probability model, \autoref{Method} outlines our strategy for developing and validating the model, \autoref{Probability} derives the probability model, \autoref{Simulation} discusses the simulation setup used, \autoref{Results} presents our results, and \autoref{Conclusion} concludes the findings of this paper.
	
	\section{Problem Formulation} \label{Problem}
	
	Our goal is to answer the following question: what is the probability that a vehicle can reach a position in a different lane at a certain distance ahead using one or multiple lane changes? In other words, what is the probability that in \autoref{Nomenclature}, the red vehicle can reach the red star? On its own, this question is not well-defined because of all the factors involved, including driving behavior, distance to the goal state, number of lanes, and state and driving behavior of nearby vehicles. Therefore, we make several simplifying assumptions to mathematically formulate the problem and develop the probability model. Later on we use traffic simulations reflecting real-world conditions to evaluate the developed model, and these assumptions are revisited when we compare model predictions with simulation results. \par
	
	Without loss of generality, we focus on a highway road segment under free-flow traffic conditions. The road segment has $n$ lanes ($n \ge 2$) numbered from left to right by 1 to $n$, i.e. the leftmost lane is lane 1 and the rightmost lane is lane $n$. We assume that vehicles in lane $i$, $1 \le i \le n$, are all passenger cars represented by a point at their center of gravity and all have the same velocity $v_{i}$ that does not change over time, but may be different from velocity $v_{j}$ where $i \neq j$. In reality, velocity varies from vehicle to vehicle and over time, so $v_{i}$ is assumed to represent the temporal average of the velocity of all vehicles traveling in lane $i$. Furthermore, we assume that headway distances (front bumper to front bumper) in lane $i$ are independent identically distributed (i.i.d.) random variables $H_{i, k}$ that have a common cumulative distribution function $F_{i}(h)$. Since our focus is on highway driving, we assume that headway distances on lane $i$ are from a log-normal probability distribution defined by parameters $(\mu_{i}, \sigma_{i})$ \cite{Mei, Yin}. Note that this choice of probability distribution does not affect the overall derivation process in \autoref{Probability} and for a different traffic condition other distributions, such as exponential, log-logistic, or Weibull, can be used as well. \par
	
	Throughout this paper we assume that the lane changing behavior of the vehicle under study (the ego vehicle) follows a Gipps gap acceptance lane-change model \cite{Gipps}, where the vehicle changes lanes only if the headway distance between its leading and trailing vehicles in the adjacent target lane $j$, $1 \le j \le n$, is larger than a minimum acceptable (critical) gap $g_{j}$. As is the case for velocity, we assume that $g_{j}$ does not change over time but can be different for different lanes. Additionally, we assume that once the vehicle finds such acceptable gap (i.e. has a distance of at least $\frac{g_{j}}{2}$ to its leading and trailing vehicles in the adjacent target lane), it instantly starts changing lanes and completes it in a set time $t_{j}$, after which its velocity matches the velocity of vehicles in the target lane $v_{j}$. Here we assume that only the ego vehicle changes lanes and none of the other vehicles do so, though statistically speaking at any instant the number of vehicles entering a lane is the same as those exiting that lane \cite{Goswami}, so our assumption should not have a large impact on the results\footnote{If there is a statistically significant imbalance between the number of vehicles entering and leaving a particular lane, it would lead to a gradual rise or decline in the vehicle density of that lane, contradicting the free-flow assumption.}. \par
	
	A near-term goal state is defined here as a point beyond the line of sight of the driver or perception sensors, but not so far that reaching it requires maneuvers beyond multiple lane changes. Denoting by $d$ the longitudinal distance (distance along the road) from the current position to the goal state, in this work we assume that $0.1\,\mathrm{km} \le d \le 5\,\mathrm{km}$. Definitions and notations above are summarized in \autoref{Nomenclature}. \par
	
	Based on these definitions, the problem can be formulated as follows: what is the success probability of reaching a point on lane $j$ a distance $d$ ahead of the current position on lane $i$, $j \neq i$, assuming that for vehicles on lane $k$, $1 \le k \le n$, all velocities are equal to $v_{k}$, headway distances are i.i.d. random variables from a log-normal distribution defined by parameters $(\mu_{k}, \sigma_{k})$, and the ego vehicle changes lanes according to the Gipps gap acceptance lane change model with critical gap $g_{k}$?
	
	\section{Methodology} \label{Method}
	
	Our work for developing the probability model starts with its derivation for the case where $n = 2$. As explained in \autoref{Problem}, the derivation starts with the assumption that the probability in question is a function of 7 parameters ($d, v_{1}, v_{2}, \mu_{2}, \sigma_{2}, g_{2}$, and $t_{2}$). Using parameter reduction techniques that exploit the relationship between these parameters, the probability is eventually formulated as a function of only three parameters. This allows us to numerically calculate the joint distribution of the probability over a range of those three parameters using Monte Carlo simulations. Having determined the probability for $n = 2$, we introduce a theorem that utilizes the law of total probability to present a recursive method for calculating the probability for cases with $n > 2$.
	
	Because of the assumptions that are made in developing the probability model in \autoref{Problem}, it is essential to validate the model and evaluate its performance under general traffic conditions. A simple validation scheme is to define a target position on a different lane of a (simulated or real) highway a distance $d$ ahead of a fixed starting point and count the number of vehicles that initiate a lane changing maneuver at the starting point and successfully reach the target position. Dividing this number by the number of vehicles that initiate the maneuver determines the success probability which can be compared with model predictions to determine its accuracy. Along with a comparison of the proposed model's predictions with previous works, the validation approach discussed above is implemented using VISSIM\textsuperscript{\textregistered} traffic simulations, where we also vary different parameters to understand how they impact the probability.
	
	\section{Probability Model} \label{Probability}
	
	To develop the probability model, we first consider the case where $n = 2$ and then expand the model to include cases where $n > 2$ using a recursive approach. Without loss of generality, assume that the ego vehicle is currently on lane 1 and the goal state is a distance $d$ ahead on lane 2. Our objective is to determine the probability of finding an acceptable gap in lane 2 and completing a lane change before the ego vehicle travels a distance $d$. Based on the assumptions in \autoref{Problem}, this probability, denoted by $P(S)$, is a function of $d, v_{1}, v_{2}, \mu_{2}, \sigma_{2}, g_{2}$, and $t_{2}$. That is, $P(S) = f_{2}(d, v_{1}, v_{2}, \mu_{2}, \sigma_{2}, g_{2}, t_{2})$ (index 2 of function $f$ refers to the number of lanes). \par
	
	The success probability $P(S)$ of reaching the goal state a distance $d$ ahead is equivalent to the probability of finding an acceptable gap in lane 2 during the time the ego vehicle travels a total distance $d_{i} = d - t_{2}v_{1}$, given that it takes $t_{2}$ seconds to change lanes. In other words, because it takes $t_{2}$ seconds to complete a lane change, the ego vehicle would miss the goal state while driving at velocity $v_{1}$ if it started changing lanes after traveling a total distance $d_{i}$. Based on the assumption that the vehicle follows a Gipps gap acceptance lane change model, starting a successful lane changing maneuver while traveling a distance $d_{i}$ is equivalent to finding an acceptable gap during that distance. \par
	
\begin{figure}[t!]
		\centering
		\includegraphics[width = \columnwidth]{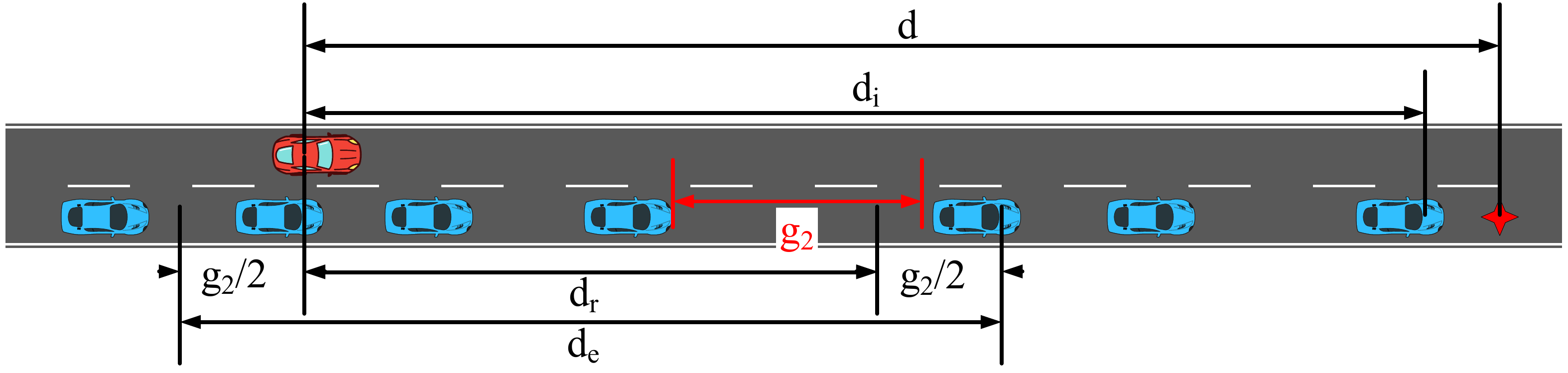}
		\caption{An illustration of parameters $d, d_{i}, d_{r}$, $d_{e}$, and their relationship to each other. Success probability $P(S)$ is equal to the probability of finding a gap in lane 2 no smaller than $g_{2}$ along $d_{e}$.} \label{ParamR}
	\end{figure}

	Denote by $d_{r}$ the distance traveled by the ego vehicle relative to lane 2 while traveling a total distance $d_{i}$. That is,
	\begin{equation} \label{RelD}
		d_{r} = d_{i}\vert\left(1 - \frac{v_{2}}{v_{1}}\right)\vert.
	\end{equation}
	The ego vehicle travels a distance $d_{r}$ relative to lane 2 while traveling a total distance $d_{i}$ and searching for an acceptable gap, so if we freeze the motion of vehicles on lane 2 and only consider the motion of the ego vehicle relative to them, $P(S)$ is equal to the probability of finding a point on lane 2 along distance $d_{r}$ ahead of (or behind if $v_{2} > v_{1}$) the current position that is at least $\frac{g_{2}}{2}$ away from each of its two nearest vehicles on that lane. The probability of finding such a point, in turn, is equivalent to the probability of finding a gap no smaller than $g_{2}$ along distance $d_{e} = d_{r} + g_{2}$ in lane 2 starting from a distance $\frac{g_{2}}{2}$ behind (or ahead of if $v_{2} > v_{1}$) the ego vehicle position. In other words, finding a point along distance $d_{r}$ that is at least $\frac{g_{2}}{2}$ away from each of its two nearest vehicles is equivalent to finding a gap no smaller than $g_{2}$ along distance $d_{e}$. This derivation is illustrated in \autoref{ParamR}.
	
	The problem of finding $P(S)$ can now be formulated in a more abstract way. Assume that the real line $\mathbb{R}$ is populated with points such that the inter-arrival distances are i.i.d. random variables from a log-normal probability distribution defined by parameter pair $(\mu_{2}, \sigma_{2})$. These points represent vehicles on lane 2 whose headway distances are i.i.d. random variables from a log-normal probability distribution. In this formulation, $P(S)$ is equal to the probability of finding a gap no smaller than $g_{2}$ in the interval $[0, d_{e}]$. Before moving on, the following lemma is needed to calculate $P(S)$.
	
	\begin{lemma} \label{LnNLemma}
		If $X$ is a log-normal random variable from a distribution with parameters ($\mu$, $\sigma$) and $k$ is a positive real number, then $Y = \frac{X}{k}$ is a log-normal random variable from a distribution with parameters ($\mu - \ln(k)$, $\sigma$).
	\end{lemma}
	\begin{IEEEproof} \label{LnNProof}
		For $y > 0$, we have
		\begin{equation} \label{LnNEq}
			\begin{split}
				P[Y \le y] &= P[\frac{X}{k} \le y] = P[X \le ky]\\
				&= \frac{1}{2} + \frac{1}{2}\mathrm{erf}[\frac{\ln(ky) - \mu}{\sqrt{2}\sigma}]\\
				&= \frac{1}{2} + \frac{1}{2}\mathrm{erf}[\frac{\ln(y) - (\mu - \ln(k))}{\sqrt{2}\sigma}],
			\end{split}
		\end{equation}
		where erf is the error function. According to (\ref{LnNEq}), $Y$ is a log-normal random variable from a distribution with parameters ($\mu - \ln(k)$, $\sigma$).
	\end{IEEEproof}
	
	Up to this point, we have reduced the number of parameters from the original 7 down to $d_{e}, g_{2}, \mu_{2}$, and $\sigma_{2}$. The next step is scaling the abstract formulation by a factor of $\frac{1}{d_{e}}$. This way, using \autoref{LnNLemma} we can deduce that $P(S)$ is equal to the probability of finding a gap no smaller than $g = \frac{g_{2}}{d_{e}}$ in the interval $[0, 1]$, assuming $\mathbb{R}$ is populated with points such that the inter-arrival distances are i.i.d. random variables from a log-normal distribution defined by parameter pair $(\mu = \mu_{2} - \ln(d_{e}), \sigma = \sigma_{2})$. This final step shows that probability $P(S)$ can be calculated based on three parameters, $g, \mu$, and $\sigma$. In other words, $P(S) = f_{2}(d, v_{1}, v_{2}, \mu_{2}, \sigma_{2}, g_{2}, t_{2}) = q(g, \mu, \sigma)$. Function $q$ can be seen as the joint distribution of the success probability over the three parameters $g, \mu$, and $\sigma$. \par
	
	Now that the number of parameters involved has been reduced from 7 to 3, a numerical approach can be used to calculate the value of $q(g, \mu, \sigma)$ for a range of tuples $(g, \mu, \sigma)$. The computations were carried out using a MATLAB\textsuperscript{\textregistered} script run on a node of Virginia Tech's NewRiver Advanced Research Computing system that has two 12-core processors \cite{NewRiver}. The probabilities were calculated for every tuple of the form $(g, \mu, \sigma)$ where $g$ varied from 0 to 1 in steps of 0.01, $\mu$ varied from -5 to 1 in steps of 0.05, and $\sigma$ varied from 0 to 2 in steps of 0.05. The resulting $101\times 121\times 41$ numerical matrix could then be used to calculate $P(S)$ for any arbitrary values of $(g, \mu, \sigma)$ through interpolation or very rarely, extrapolation. The above range of values for parameters $g, \mu$, and $\sigma$ were chosen to cover most, if not all, traffic scenarios on the road. For example, in a scenario where $\mu$ is 1 the average headway along lane 2 is $ed_{e}$ in the original abstract formulation. Alternatively, in a scenario where $\mu$ is -5 the average headway along lane 2 is around 0.00674$d_{e}$. For context, given that $d$ is limited to 5 km and vehicles travel along a highway, $d_{e}$ hardly rises beyond 1 km, meaning the average headway along lane 2 for this extreme case is no more than 7 meters, barely longer than the length of a vehicle.
	
	\setlength{\tabcolsep}{2.9 pt}
	\begin{table}[t!]
		\renewcommand{\arraystretch}{1.3}
		\setlength{\abovecaptionskip}{2 pt}
		\caption{Probability $P(S)$ for Different Values of $g, \mu$, and $\sigma$} \label{SimNum}
		\centering
		\begin{tabular}{c c c | c c c c c c c | c}
			\hline
			\multicolumn{3}{c}{$P(S)$} & \multicolumn{7}{c}{$\log_{10}(N)$} & Absolute \\
			\cline{1 - 10}
			$g$ & $\mu$ & $\sigma$ & 1 & 2 & 3 & 4 & 5 & 6 & 7 & error (\%) \\
			\hline
			0.2 & -2 & 0.4 & 0.6 & 0.63 & 0.698 & 0.7029 & 0.6928 & 0.6921 & 0.6924 & 0.04 \\
			0.2 & -2 & 0.8 & 1 & 0.96 & 0.966 & 0.9569 & 0.9549 & 0.9537 & 0.9538 & 0.11\\
			0.2 & -1 & 0.4 & 1 & 1 & 1 & 1 & 1 & 1 & 1 & 0 \\
			0.2 & -1 & 0.8 & 1 & 1 & 0.999 & 0.9999 & 0.9998 & 0.9999 & 0.9999 & 0.01 \\
			0.5 & -2 & 0.4 & 0 & 0 & 0.002 & 0.0018 & 0.0023 & 0.0021 & 0.0021 & 0.02 \\
			0.5 & -2 & 0.8 & 0.3 & 0.20 & 0.217 & 0.2056 & 0.2019 & 0.2021 & 0.2012 & 0.07 \\
			0.5 & -1 & 0.4 & 0.1 & 0.32 & 0.382 & 0.3643 & 0.3545 & 0.3564 & 0.3567 & 0.22 \\
			0.5 & -1 & 0.8 & 0.7 & 0.72 & 0.658 & 0.6597 & 0.6584 & 0.6600 & 0.6602 & 0.18 \\
			\hline
		\end{tabular}
	\end{table}
	\setlength{\tabcolsep}{6 pt}
	\setlength{\abovecaptionskip}{0.5\baselineskip}
	
	For each tuple of the form $(g, \mu, \sigma)$, probability $P(S) = q(g, \mu, \sigma)$ was calculated through numerical Monte Carlo simulation of the abstract problem. Specifically, the script first generated a large number of random values from a log-normal distribution defined by parameters $\mu$ and $\sigma$, representing the inter-arrival distances. Then, it calculated the cumulative sum of those random values which represent the placement of points on $\mathbb{R}$ in the abstract formulation. In the next step, the script selected a unit interval within the bounds of the cumulative sum at random and checked for a gap larger than or equal to $g$ in that interval. This process was repeated $10 ^ {5}$ times for any given value of parameters $g, \mu$, and $\sigma$, and the value of $q(g, \mu, \sigma)$ was determined by dividing the number of successful cases (i.e. cases where an acceptable gap was found) by $10 ^ {5}$. This approach is supported by the law of large numbers in probability theory, which states that as the number of iterations increases, the calculated numerical probability converges to the real value \cite{Leon}. \autoref{SimNum} shows that for random values of $g, \mu$, and $\sigma$, as the number of iterations $N$ is increased from 10 to $10 ^ {7}$, the probability converges and for $10 ^ {5}$ iterations, the calculated value is accurate to at least 2 decimal points compared to the one calculated for $10 ^ {7}$ iterations. \par
	
	Now that we can calculate the value of $P(S)$ for $n = 2$, i.e. $f_{2}(d, v_{1}, v_{2}, \mu_{2}, \sigma_{2}, g_{2}, t_{2})$, its value for cases with $n > 2$ can be calculated using the following theorem.
	
	\begin{theorem} \label{RecursiveThm}
		For $n \ge 3$, probability $P(S) = f_{n}(d, v_{1 : n}, \mu_{2 : n}, \sigma_{2 : n}, g_{2 : n}, t_{2 : n})$ (where $w_{l : m}$ means $w_{1}, w_{2},\ldots, w_{m}$ for any parameter $w$ and indices $l \le m$) is calculated by the following equation.
		\begin{equation} \label{RecursiveEqn}
			\begin{split}
				&f_{n}(d, v_{1 : n}, \mu_{2 : n}, \sigma_{2 : n}, g_{2 : n}, t_{2 : n})\\
				&= \int_{0} ^ {d} f_{2}(d - x, v_{n - 1 : n}, \mu_{n}, \sigma_{n}, g_{n}, t_{n})\\
				&\times\frac{\partial}{\partial x} f_{n - 1}(x, v_{1 : n - 1}, \mu_{2 : n - 1}, \sigma_{2 : n - 1}, g_{2 : n - 1}, t_{2 : n - 1})\mathrm{d}x \\
				&= \frac{\partial}{\partial x}\int_{0} ^ {d} f_{2}(d - x, v_{n - 1 : n}, \mu_{n}, \sigma_{n}, g_{n}, t_{n})\\
				&\times f_{n - 1}(x, v_{1 : n - 1}, \mu_{2 : n - 1}, \sigma_{2 : n - 1}, g_{2 : n - 1}, t_{2 : n - 1})\mathrm{d}x.
			\end{split}
		\end{equation}
	\end{theorem}
	\begin{IEEEproof} \label{RecursiveProof}
		Using the continuous form of the law of total probability \cite{Leon}, one can write
		\begin{equation} \label{TotalProb}
			P(S) = \int_{-\infty} ^ {\infty} P(S\vert X = x)f_{X}(x)\mathrm{d}x.
		\end{equation}
		Here, conditioning the probability on location $x$ where the vehicle changes lanes from lane $n - 2$ to lane $n - 1$, $0 \le x \le d$, it is easy to see that $P(S\vert X = x)$ is nothing but $f_{2}(d - x, v_{n - 1 : n}, \mu_{n}, \sigma_{n}, g_{n}, t_{n})$, because it is the probability of successfully reaching the goal state on lane $n$ a distance $d - x$ ahead from the current position $x$ on lane $n - 1$. As for $f_{X}(x)$, note that it is the derivative of $F_{X}(x)$, the cumulative distribution function of the probability that at some point on or before $x$ the vehicle reached lane $n - 1$ using $n - 2$ lane changes. This probability, in turn, is equal to the success probability for a case where the goal state is located a distance $x$ ahead on lane $n - 1$. Therefore, $f_{X}(x) = \frac{\partial}{\partial x} f_{n - 1}(x, v_{1 : n - 1}, \mu_{2 : n - 1}, \sigma_{2 : n - 1}, g_{2 : n - 1}, t_{2 : n - 1})$. Substituting these values in (\ref{TotalProb}) and limiting the integration bounds to 0 and $d$ gives (\ref{RecursiveEqn}), completing the proof.
	\end{IEEEproof}
	
	\autoref{RecursiveThm} shows that in a real-world implementation, the system only needs the (calculated or estimated) parameter values specified above to estimate the success probability. For example, if a vehicle is on the leftmost lane of a three-lane highway and needs to reach a target position on the rightmost lane a distance $d$ ahead, calculation of the success probability requires knowledge of the values of $d, v_{1}, v_{2}, v_{3}, \mu_{2}, \mu_{3}, \sigma_{2}, \sigma_{3}, g_{2}, g_{3}, t_{2}$, and $t_{3}$. These values can be estimated once and provided to the system, or can be updated periodically to improve the probability estimate. When the vehicle moves to lane 2, calculations now only require $d ^ {\prime}$ (new distance to the target position), $v_{2}, v_{3}, \mu_{3}, \sigma_{3}, g_{3}$, and $t_{3}$ to estimate the probability. More details about different approaches to implementing the probability model are available in \cite{Mehr2, Mehr3, Mehr4}. \par
	
	To summarize, the calculated numerical matrix for the base case along with \autoref{RecursiveThm} allow one to calculate the probability of reaching a goal state on a road segment with $n$ lanes using one or multiple lane changes.
	
	\section{Simulation Setup} \label{Simulation}
	
	Despite the simple scheme outlined in \autoref{Method}, validation of the developed model through experiments or simulation is inherently challenging. This is because as soon as a vehicle has the intent to change (one or multiple) lanes to reach a goal state, that intent can affect driving behavior - for example by slowing down or pursuing an aggressive lane changing strategy - and this deviation from the initial conditions can result in an outcome that is not representative of the true success probability of the initial state that the model is trying to estimate, introducing error into the validation process. Assume, for example, that the ego vehicle is driving in lane 1 at speed $v$ and the goal state is a distance $d$ ahead on lane 2. To obtain the true success probability $P_{t}$ of reaching the goal state that the model is trying to estimate, the ego vehicle has to keep driving at speed $v$ and see how likely it is to find an acceptable gap in and move to lane 2. If this process is repeated 100 times and in 80 of those the ego vehicle finds an acceptable gap, then $P_{t} = 80\%$. On the other hand, if the ego vehicle slows down during this process - as most drivers do when they want to change lanes - that changes the likelihood of finding an acceptable gap in lane 2. Consequently, if we measure the new success probability $P ^ {\prime}_{t}$ it will be different from the true success probability $P_{t}$ that we intended to measure, corrupting the validation process. Therefore, it is essential that each vehicle maintain its driving behavior. This criterion, along with the fact that an experimental validation process requires a large set of trials for any single case which is both time-consuming and overly costly, makes experimental validation infeasible. Therefore, we opted for using PTV VISSIM\textsuperscript{\textregistered} 11.00-12 to simulate traffic flow and validate the probability model \cite{PTV}.
	
	\begin{table}[t!]
		\renewcommand{\arraystretch}{1.3}
		\setlength{\abovecaptionskip}{4 pt}
		\caption{Simulation Computer Specifications} \label{Comp}
		\centering
		\begin{tabular}{c c}
			\hline
			Component & Specification \\
			\hline
			CPU & Intel\textsuperscript{\textregistered} Core-i7 6700HQ @ 3.13 GHz \\
			RAM & 16 GB @ 2133 MHz \\
			\multirow{2}{*}{GPU} & Intel\textsuperscript{\textregistered} HD Graphics 530 \\
			 & Nvidia\textsuperscript{\textregistered} GeForce GTX 960M \\
			\hline
		\end{tabular}
	\end{table}
	\setlength{\abovecaptionskip}{0.5\baselineskip}
	
	Simulations were carried out on a computer with specifications listed in \autoref{Comp}. They were performed on a 10 km road segment with one input and one output that was modeled as a single link, with each simulation running for 72000 seconds. Depending on the case being simulated, the road segment had either 2, 3, or 4 lanes. Given our focus on highway driving and based on the recommendation of \cite{VissimGuide}, the default Freeway (free lane selection) driving behavior - using the Wiedemann 99 car following model \cite{Wiedemann} with default parameter values - was applied to the road segment. Finally, desired velocity along each lane was set using sets of Desired Speed Decision points. For each vehicle passing through such a point with nominal desired velocity $V$, a random desired velocity in the interval $V \pm 5$ km/h with Gaussian probability density would be assigned to the vehicle. We should note that these points only set the desired velocity, i.e. velocity under free flow traffic conditions (you can think of them as the equivalent of speed limits in the real world), and the actual velocity of a vehicle could vary and be significantly lower depending on the traffic condition. \par
	
	Two classes of vehicles were used in the simulation. The first class (here called normal vehicles) consisted of vehicles that were completely controlled by VISSIM\textsuperscript{\textregistered}'s internal model (Wiedemann 99 \cite{Wiedemann}). The second class (here called target vehicles) consisted of vehicles identical to the first class, with the only difference in that their lane change initiation behavior was controlled through VISSIM\textsuperscript{\textregistered}'s external driver model (EDM) API, which grants the user control over various aspects of the driving behavior of all or a group of vehicles. The EDM forced each target vehicle that entered the road to change lanes and move to and stay in the leftmost lane for the first 5 kilometers. At the 5 km mark, the EDM instructed each target vehicle to change lanes until it reached the rightmost lane. Starting from the 5 km mark, vehicle counters were placed at 500 m intervals up to the 10 km mark to count the number of target vehicles passing the 5 km mark in the leftmost lane and those passing every other counter in the rightmost lane. These values could then be used to calculate the probability along the 5 km distance. \par
	
	At the road entrance, normal and target vehicle fractions were set to 0.98 and 0.02, respectively. This ensured that target vehicle behavior had a small impact on the overall flow of traffic while at the same time generating a large enough number of target vehicles during a 20 hour simulation to be statistically significant, helping increase the accuracy of the measured probability. Furthermore, the EDM was programmed to instruct each target vehicle to change lanes after the 5 km mark only when that vehicle's longitudinal (along the road) distance from its leading and trailing vehicles on the adjacent lane to the right was at least $\frac{s_{0}}{2} + \frac{\delta}{2}v_{l}$ and $\frac{s_{0}}{2} + \frac{\delta}{2}v_{t}$, respectively. Here, $v_{l}$ and $v_{t}$ denote the velocity of the leading and trailing vehicles, respectively, $s_{0}$ is a constant, and $\delta$ is the minimum desired time headway between vehicles on the adjacent lane. In other words, target vehicles considered a $\frac{\delta}{2}$ safety zone (time headway) around the leading and trailing vehicles on the adjacent lane and changed lanes only when the two zones were non-overlapping \cite{Cesari, Chandra}. As such, target vehicles only changed lanes when the gap to their right was at least $g = s_{0} + \delta v_{m}$, where $v_{m} = \frac{v_{l} + v_{t}}{2}$. In the probability model, since $g_{i}$ for each lane $i$, $2 \le i \le n$, was assumed to be constant over time, it can be assumed that $g_{i} = s_{0} + \delta v_{i}$, where $v_{i}$ is the spacial and temporal average velocity of lane $i$ during the simulation. For our simulations, $s_{0}$ was always set to 7 m. Finally, VISSIM\textsuperscript{\textregistered} completes a lane change in 3 seconds from when it is initiated; therefore, we set $t_{i} = 3\,\mathrm{s},\, 2 \le i \le n$, in our model. \par
	
	Simulation data was recorded and processed after each run. It included the number of target vehicles that passed through the counters, spacial and temporal average of velocities of all vehicles on each lane, and headway distance distribution along each lane over the course of the simulation. Headway distance distribution was used to estimate the value of parameters $\mu$ and $\sigma$ for each lane. These values were entered into the probability model along with average velocity values to estimate probability $P(S)$ along the 5 km distance. Separately, data from the counters was used to determine probability $P(S)$ at 500 meter intervals along the 5 km distance. \par
	
	As discussed in \autoref{Probability}, several parameters influence probability $P(S)$, including average velocity $v$ of each lane, overall traffic density per lane $\rho_{l}$, target distance $d$, the number of lanes $n$, and driving behavior. This results in a large parameter space which makes simulating every possible combination of parameter values to validate the probability model impossible. Therefore, our validation strategy was to start from a base case and vary parameters one by one in a certain interval, comparing model predictions with simulation results at each step. This approach helped both validate the probability model and shed light on the effects of different parameters on $P(S)$. For our base case, $\rho_{l}$ was set to 1200 veh/h/ln, $\delta$ was set to 2 s, and the desired velocity was set to 100 km/h for the rightmost lane, increasing by 10 km/h for each lane to the left. All simulation cases were repeated for 2, 3, and 4 lanes, so for example in the base case for three lanes, traffic density $\rho$ was 3600 veh/h and the desired velocity of the rightmost, middle, and leftmost lanes were 100 km/h, 110 km/h, and 120 km/h, respectively. \par
	
	Simulation results in \autoref{Results} cover the effects of five different parameters on probability $P(S)$, namely $\rho_{l}$, $\delta$, $v_{1}$, $n$, and $d$. Details of the values used for each parameter are shown in \autoref{ParamRange}. In total, 51 simulations were carried out, 17 for each value of $n$.
	
	\begin{table}[t!]
		\renewcommand{\arraystretch}{1.3}
		\caption{Range of Simulation Parameters} \label{ParamRange}
		\centering
		\begin{tabular}{c c c c c}
			\hline
			Parameter & Unit & Base value & Range & Step \\
			\hline
			$\rho_{l}$ & veh/h/ln & 1200 & 400 - 2400 & 400 \\
			$\delta$ & s & 2.0 & 0.4 - 3.2 & 0.4 \\
			$v_{1} - v_{2}$ & km/h & 10 & -20 - 20 & 10 \\
			$n$ & - & - & 2 - 4 & 1 \\
			$d$ & km & - & 0 - 5 & - \\
			\hline
		\end{tabular}
	\end{table}
	
	\section{Results and Discussion} \label{Results}
	
	Up to this point, we formulated the problem in \autoref{Problem}, developed the probability model in \autoref{Probability}, and discussed the simulation setup for validating the model in \autoref{Simulation}. \autoref{Vis} presents numerical probabilities calculated for the abstract problem in \autoref{Probability}. Then, \autoref{Validation} discusses and compares model predicitions and VISSIM\textsuperscript{\textregistered} simulation results.
	
	\subsection{Probability matrix visualization} \label{Vis}
	
\begin{figure}[t!]
		\centering
		\includegraphics[width = \columnwidth]{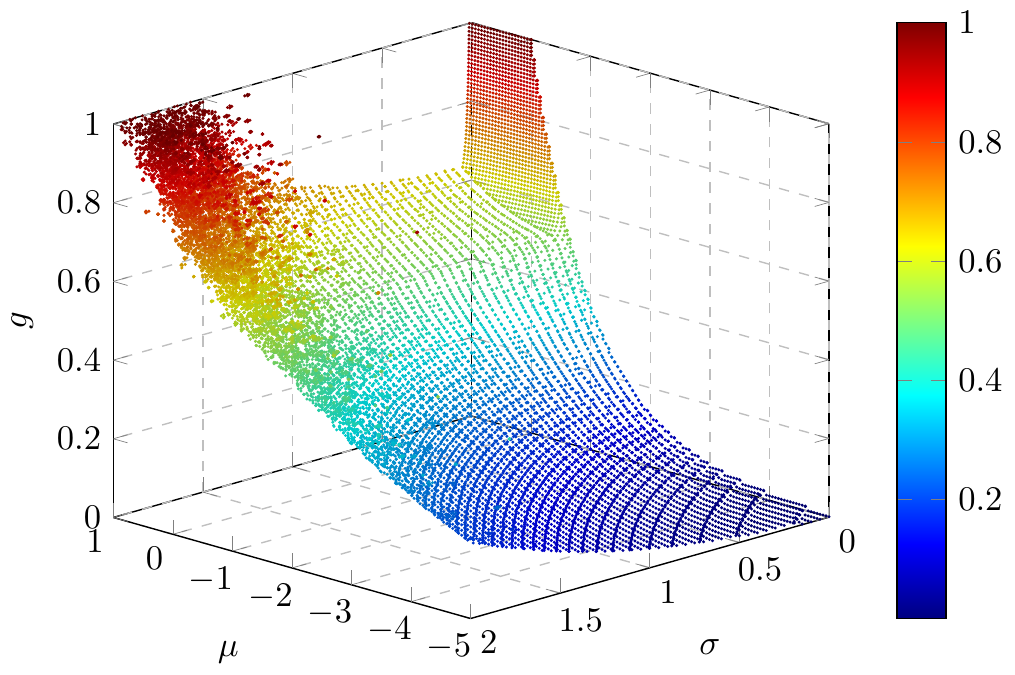}
		\caption{The isosurface representing tuples $(g, \mu, \sigma)$ for which $q(g, \mu, \sigma) = 0.9$. For better visualization, points are colored according to the value of $g$.} \label{NumVis}
	\end{figure}

	In \autoref{Probability} we showed that for cases with only two lanes $P(S) = f_{2}(d, v_{1}, v_{2}, \mu_{2}, \sigma_{2}, g_{2}, t_{2}) = q(g, \mu, \sigma)$ and then proceeded to calculate $q(g, \mu, \sigma)$ for a range of parameter values, obtaining a 101$\times$121$\times$41 numerical matrix. To gain insight into and visualize some of these values, \autoref{NumVis} shows an isosurface of all tuples $(g, \mu, \sigma)$ for which $q(g, \mu, \sigma) = 0.9$. Similar isosurface plots can be created for different values of $q(g, \mu, \sigma)$ between 0 and 1. For better visualization, points are colored according to the value of $g$.
	
	\autoref{NumVis} shows that in general, as the value of either $\mu$ or $\sigma$ increases, the value of $g$ corresponding to $q(g, \mu, \sigma) = 0.9$ increases. This is expected for a unit interval, since by increasing either $\mu$ or $\sigma$ the random points generated move further apart from each other and the chances of finding a gap $g$ in the unit interval increases. In other words, for a larger gap $g$ one can expect $q(g, \mu, \sigma) = 0.9$. The only exception is when $\mu$ is very large (here $\mu > 0$) and $\sigma$ is very small (here $\sigma < 0.05$). In this case, inter-arrival distances are almost identical (because of the small $\sigma$) and all larger than the unit interval (because $\mu > 0$), so one can expect to find large gaps in the unit interval. In other words, for values of $g$ close to 1, $q(g, \mu, \sigma) = 0.9$. As the value of $\sigma$ increases, the uniformity of inter-arrival distances decreases and so does the value of gap $g$ for which $q(g, \mu, \sigma) = 0.9$. Finally, \autoref{NumVis} shows that for large values of $\sigma$, the value of $g$ for which $q(g, \mu, \sigma) = 0.9$ becomes increasingly unreliable due to the large variance. This is more noticeable for larger values of $\mu$.
	
	\subsection{Model validation and parameter study} \label{Validation}
	
\begin{figure}[t!]
		\centering
		\includegraphics[width = \columnwidth]{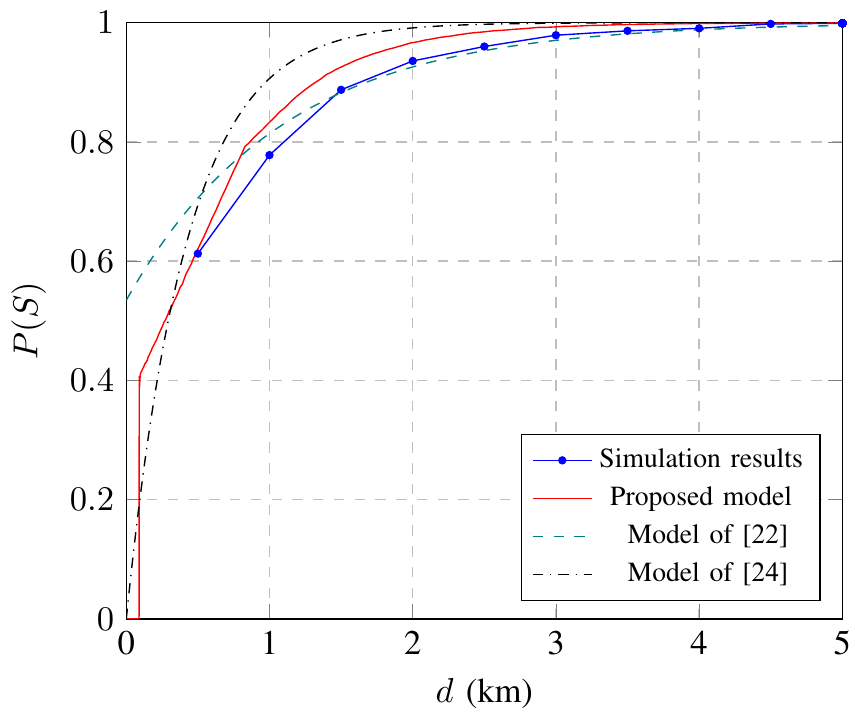}
		\setlength{\abovecaptionskip}{-10 pt}
		\caption{Comparison of the predictions of our model, the model proposed by \cite{Gong}, and the model proposed by \cite{He} with simulation results for the 2 lanes base case.} \label{Comparison}
	\end{figure}

\begin{figure}[ht!]
		\centering
		\subfloat[2 lanes]{
			\includegraphics[width = \columnwidth]{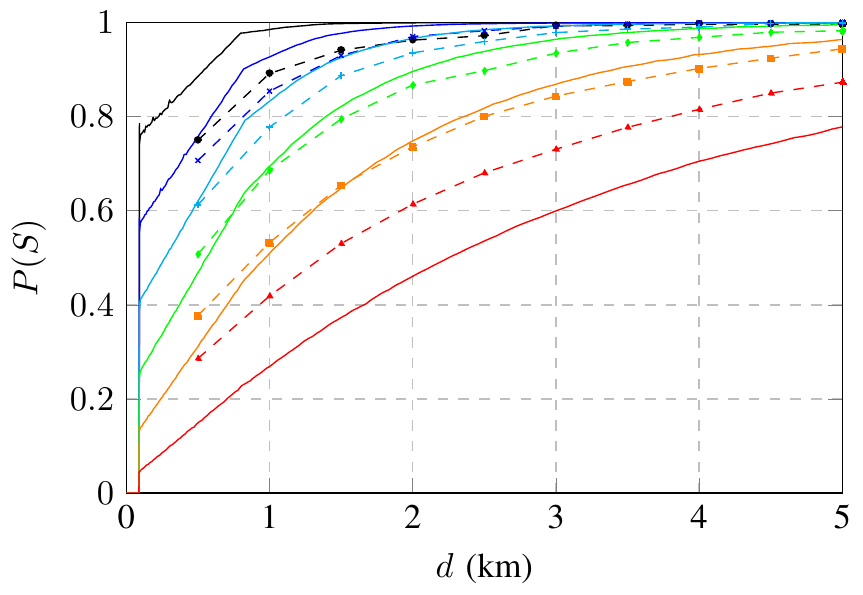} \label{Traffic2}
		} \hfill
		\subfloat[3 lanes]{
			\includegraphics[width = \columnwidth]{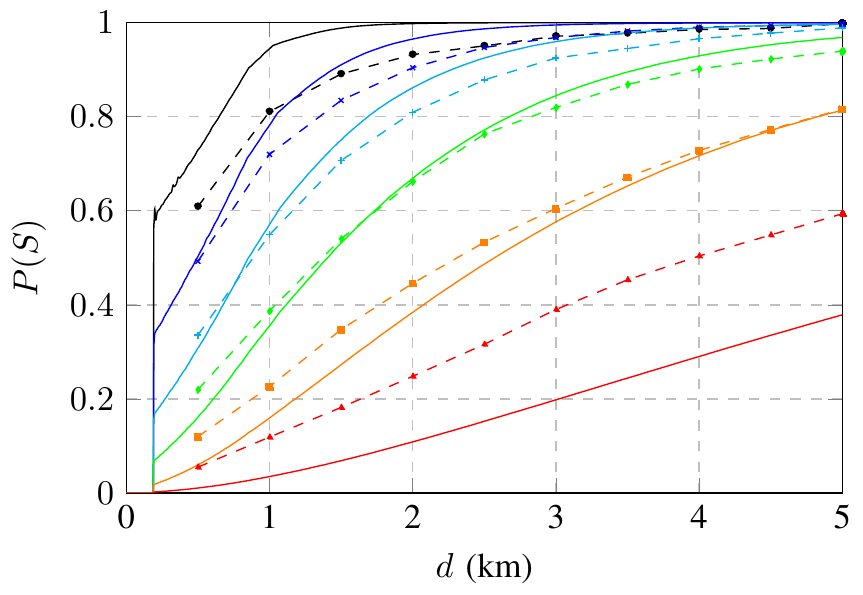} \label{Traffic3}
		} \hfill
		\subfloat[4 lanes]{
			\includegraphics[width = \columnwidth]{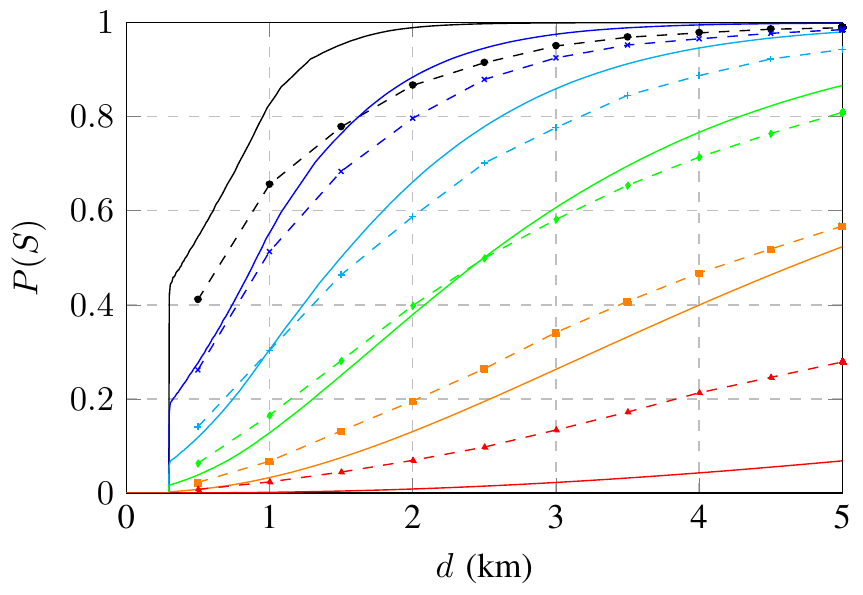} \label{Traffic4}
		} \hfill
		\subfloat{
			\includegraphics[width = 0.72\columnwidth]{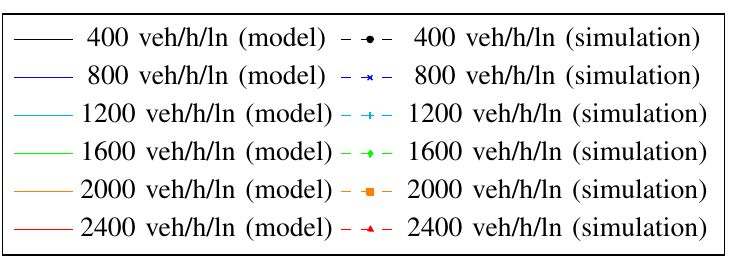}
		}
		\setlength{\belowcaptionskip}{-24 pt}
		\caption{Success probability $P(S)$ along a 5 km distance for different values of traffic density per lane $\rho_{l}$. The most deviation of model predictions from simulation results occur at limiting cases with sparse or dense traffic. For all cases, $\delta =$ 2 s.} \label{Traffic}
	\end{figure}

	In \autoref{Probability} we proposed a model to calculate the probability $P(S)$ that a vehicle reaches a near-term goal state using one or multiple lane changes and later described the simulation setup that was used to validate that probability model and investigate the effect of different parameters on $P(S)$. \autoref{Comparison} shows a comparison of the predictions of our model, the model proposed by \cite{Gong} (Model A), and the model proposed by \cite{He} (model B) with simulation results for the 2 lanes base case. Moreover, \autoref{Traffic} to \autoref{Velocity} illustrate a comparison between model predictions and simulation results while varying $\rho_{l}$, $\delta$, and $v_{1}$ from the base case, respectively.
	
	\setlength{\abovecaptionskip}{0.5\baselineskip}

	\setlength{\belowcaptionskip}{0 pt}
	\begin{figure}[ht!]
		\centering
		\subfloat[2 lanes]{
			\includegraphics[width = 0.98\columnwidth]{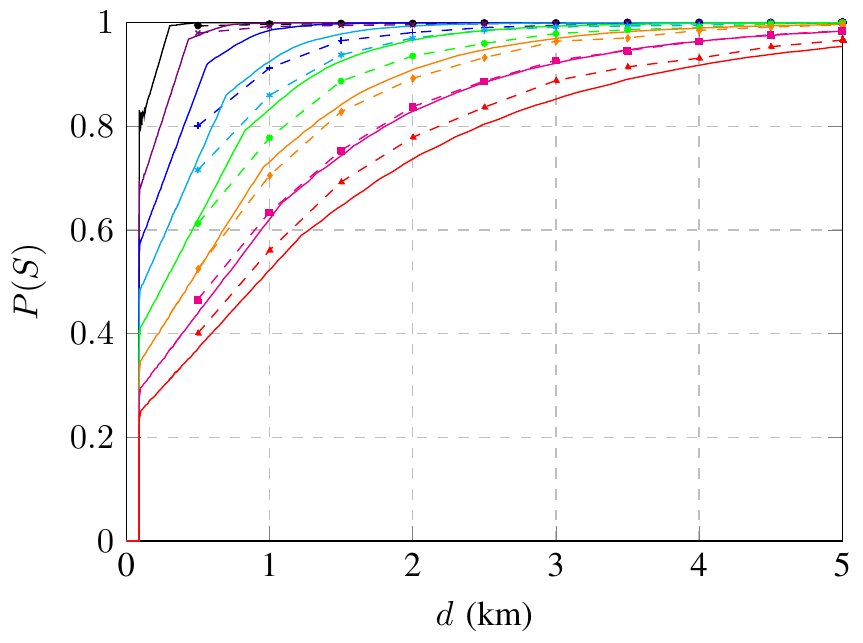} \label{Driving2}
		} \hfill
		\subfloat[3 lanes]{
			\includegraphics[width = 0.98\columnwidth]{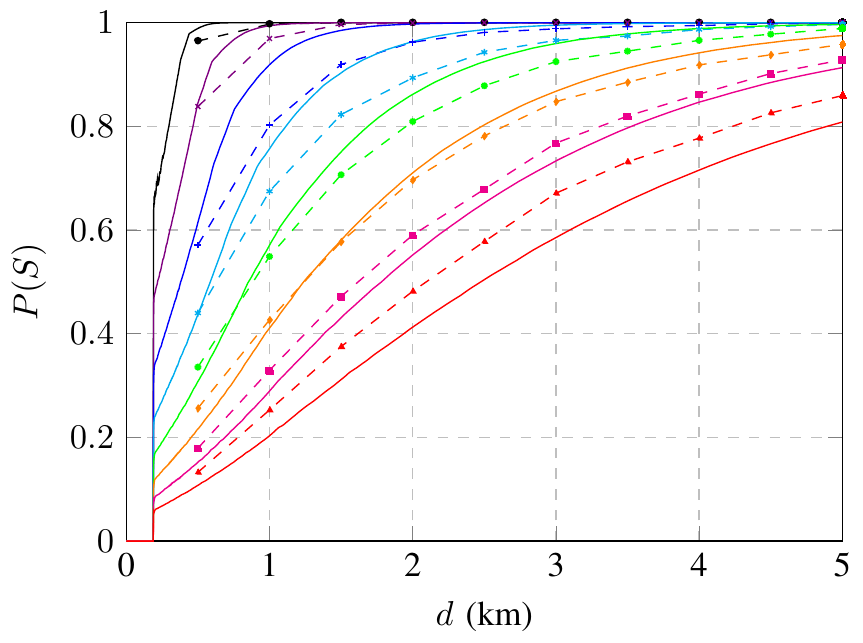} \label{Driving3}
		} \hfill
		\subfloat[4 lanes]{
			\includegraphics[width = 0.98\columnwidth]{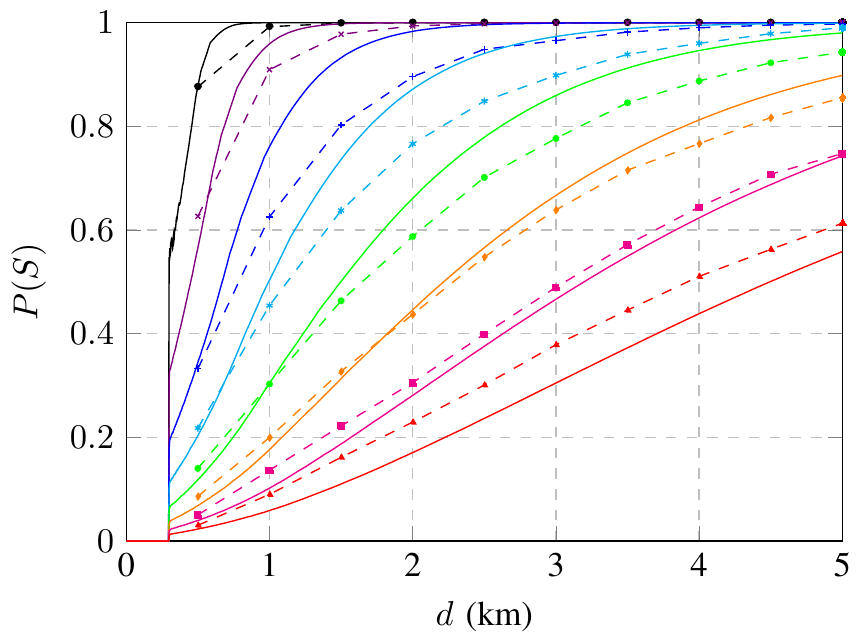} \label{Driving4}
		} \hfill
		\subfloat{
			\includegraphics[width = \columnwidth]{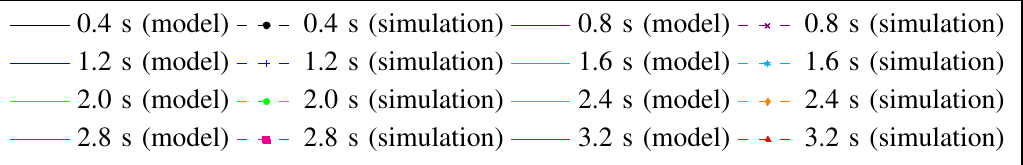}
		}
		\setlength{\belowcaptionskip}{-24 pt}
		\caption{Success probability $P(S)$ along a 5 km distance for different values of desired time headway $\delta$. For all cases, $\rho_{l} =$ 1200 veh/h/ln.} \label{Driving}
	\end{figure}
	\setlength{\belowcaptionskip}{0 pt}
	\begin{figure}[ht!]
		\centering
		\subfloat[2 lanes]{
			\includegraphics[width = \columnwidth]{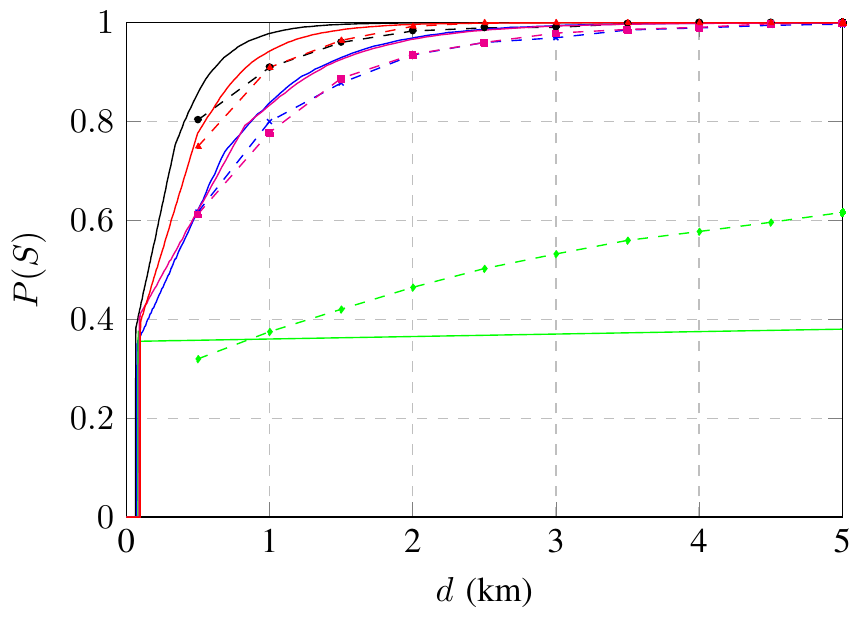} \label{Velocity2}
		} \hfill
		\subfloat[3 lanes]{
			\includegraphics[width = \columnwidth]{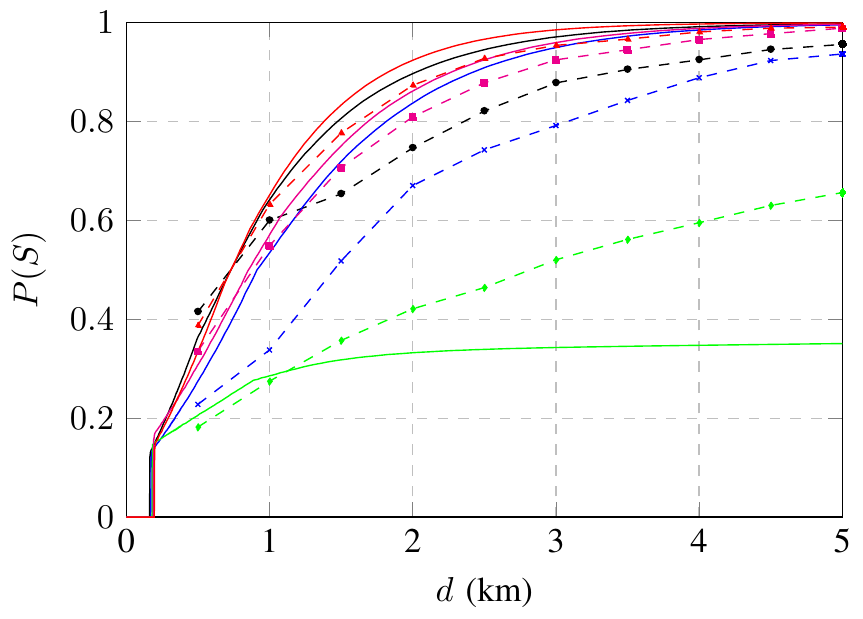} \label{Velocity3}
		} \hfill
		\subfloat[4 lanes]{
			\includegraphics[width = \columnwidth]{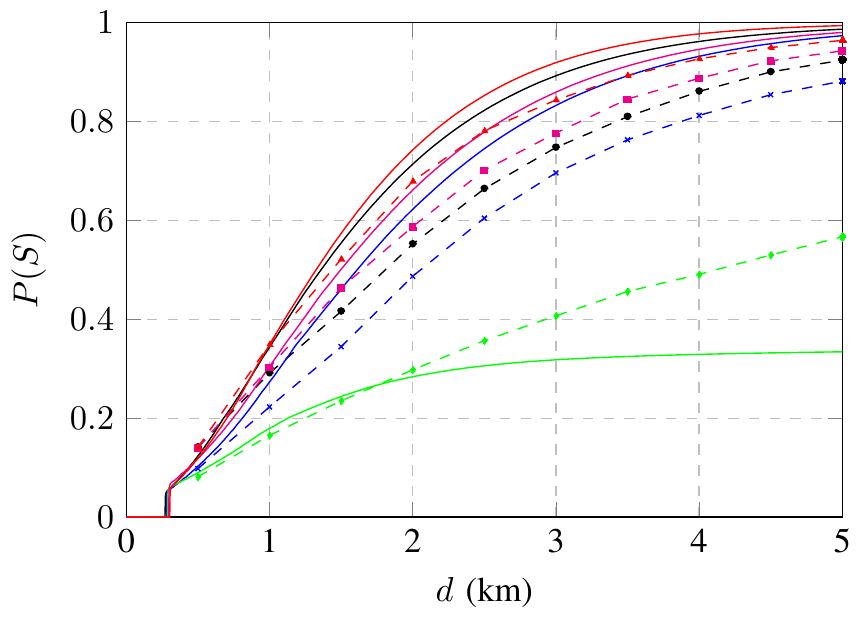} \label{Velocity4}
		} \hfill
		\subfloat{
			\includegraphics[width = 0.72\columnwidth]{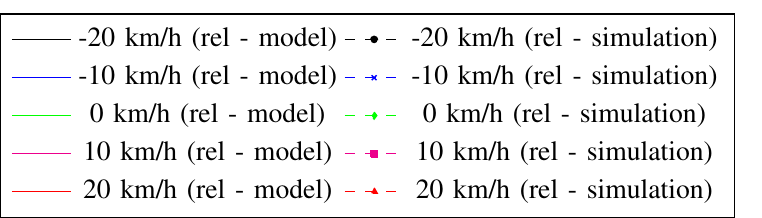}
		}
		\setlength{\abovecaptionskip}{4 pt}
		\setlength{\belowcaptionskip}{-24 pt}
		\caption{Success probability $P(S)$ along a 5 km distance for different values of average velocity $v_{1}$ relative to a constant $v_{2}$. For all cases, $\rho_{l} =$ 1200 veh/h/ln and $\delta =$ 2 s.} \label{Velocity}
	\end{figure}
	\setlength{\belowcaptionskip}{0 pt}
	\setlength{\abovecaptionskip}{0.5\baselineskip}
	
	\autoref{Comparison} shows that compared to Model B, which far overestimates the probability, our model is more accurate along the entire distance $d$. However, it is not as accurate as Model A for large values of $d$ (greater than $\sim$800 m), though the difference is small. Although none of these models are perfect, because the probability for cases with more lanes is calculated using a convolution integral and the large error for small $d$ values present in Model A can skew the results, our model seems most suitable for generalization to cases with more lanes.
	
	An overall look at \autoref{Traffic} to \autoref{Velocity} reveals that for most cases, predicted success probability $P(S)$ is accurate to within 4 percentage points. It also shows that $P(S)$ increases with distance $d$, since as $d$ increases the vehicle has more time to look for an acceptable gap in the adjacent lane and change lanes. $P(S)$, however, is not linear in terms of $d$ and can be best described using a logistics function. Furthermore, these figures show that for a fixed value of $d$ and all other parameters, $P(S)$ decreases as the number of lanes increases, though this is not linear either. It should be noted that for a small segment at the start of each plot $P(S)$ is zero because of the 3-second time it takes to change lanes ($t$). The step in $P(S)$ immediately afterwards corresponds to the probability that the vehicle could start an uninterrupted lane changing maneuver at the moment it has the intent to do so (in the simulation, at the moment it passes the 5 km mark). For example, for cases with 2 lanes, the first 90 meters correspond to the 3-second time it takes to change lanes while traveling at 30 m/s and the step in $P(S)$ after that point indicates the probability of the vehicle finding an acceptable gap right next to it at the moment it has the intent to change lanes. \par
	
\begin{figure}[t!]
		\centering
		\includegraphics[width = \columnwidth]{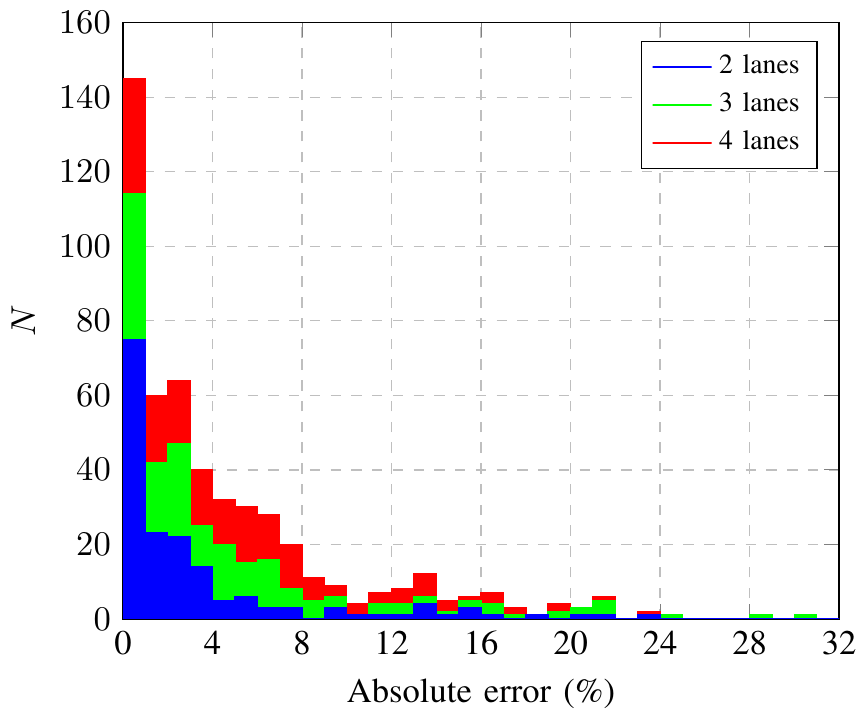}
		\setlength{\abovecaptionskip}{-16 pt}
		\caption{Histogram of the absolute error between model predictions and simulation results for all cases with 2, 3, or 4 lanes.} \label{Error}
	\end{figure}

	\autoref{Error} shows a histogram of the absolute error between model predictions and simulation results. Overall, as the number of lanes increases, model accuracy decreases. On average, the absolute error for a case with 2, 3, and 4 lanes is 3.13\%, 5.39\%, and 5.50\%, respectively.
	
	\setlength{\abovecaptionskip}{0.5\baselineskip}
	
	The errors, for the most part, are a result of the simplifying assumptions made to develop the probability model. For example, in the model we assumed that the velocity of all vehicles on lane $i$, $1 \le i \le n$, is constant over time and equal to $v_{i}$. We further assumed that headway distances are i.i.d. random variables from a log-normal probability distribution, and that only the ego vehicle changes lanes. In the simulations (as in reality), velocity varies from vehicle to vehicle and over time, headway distances have a distribution that may be different from a log-normal distribution, and other vehicles change lanes as well. A good example showcasing the differences is the case where the average velocity of both lanes is the same (0 km/h relative average velocity between lanes 1 and 2) in \autoref{Velocity2}. According to the model, since on average the ego vehicle in lane 1 does not move relative to lane 2, the success probability $P(S)$ should be constant along distance $d$ and equal to the probability of finding an acceptable gap right next to the ego vehicle. In the simulation, however, individual vehicles in each lane have different velocities and move relative to those in the other lane. Therefore, the ego vehicle has an increased chance of finding an acceptable gap further downstream, which results in $P(S)$ rising with $d$ in the simulation. \par
	
	A second, smaller source of error is the way VISSIM\textsuperscript{\textregistered} handles the EDM for target vehicles. During a simulation, although target vehicles initiate lane changing maneuvers according to the EDM, they can still receive and act on suggestions from VISSIM\textsuperscript{\textregistered}'s internal model which on some occasions may be in conflict with commands from the EDM. For example, in a case with 3 lanes where a target vehicle is in the middle lane past the 5 km mark and looking for an acceptable gap in the right lane, because the left lane has a higher average velocity, VISSIM\textsuperscript{\textregistered}'s internal model may decide to instruct the vehicle to move to the left lane, resulting in momentary zigzag behavior. For cases with more than 2 lanes, this model conflict may slightly skew simulation results. \par
	
	\begin{figure}[t!]
		\centering
		\includegraphics[width = \columnwidth]{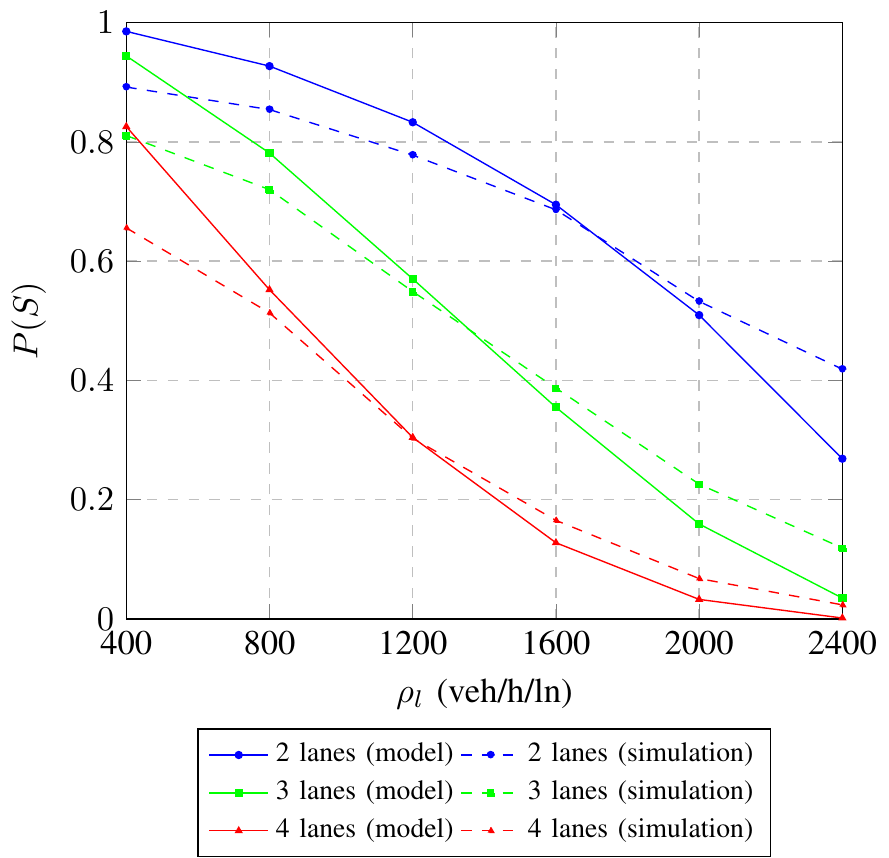}
		\caption{Success probability $P(S)$ as a function of traffic density per lane $\rho_{l}$ at a distance $d$ of 1 km.} \label{TrafficComp}
	\end{figure}

		\autoref{Traffic} shows success probability $P(S)$ along a 5 km distance for different values of traffic density per lane $\rho_{l}$ for cases with 2, 3, and 4 lanes. As $\rho_{l}$ increases, $P(S)$ decreases since the gaps between vehicles in each lane shrink and that reduces the probability of finding an acceptable gap. This reduction is not linear though, as shown in \autoref{TrafficComp} for $P(S)$ values at $d =$ 1 km, and follows an S-shaped curve. The most deviation of model predictions from simulation results occur at limiting cases where $\rho_{l}$ is either small or large, corresponding to sparse or dense traffic. To put it in context, for $\rho_{l}$ = 400 veh/h/ln the average time headway between consecutive vehicles is 9 seconds while for $\rho_{l}$ = 2400 veh/h/ln it is only 1.5 seconds. \par
	
	\begin{figure}[t!]
		\centering
		\includegraphics[width = \columnwidth]{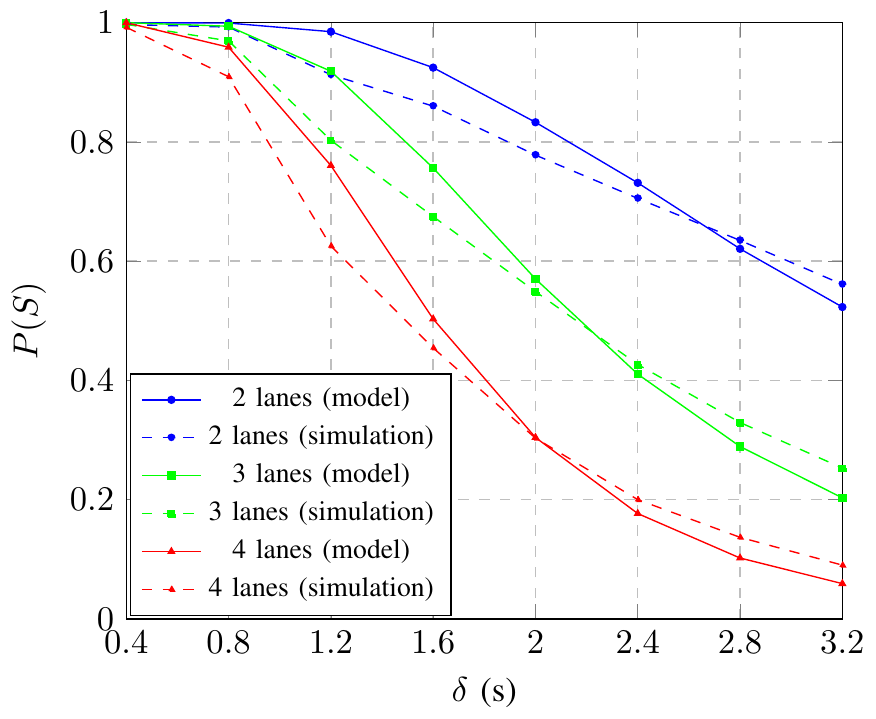}
		\caption{Success probability $P(S)$ as a function of desired time headway $\delta$ at a distance $d$ of 1 km.} \label{DrivingComp}
	\end{figure}

	Effects of $\delta$ on $P(S)$ are shown in \autoref{Driving} along a 5 km distance for cases with 2, 3, and 4 lanes. Small values of $\delta$ correspond to an aggressive driving behavior where the vehicle changes lanes upon finding very small gaps in the adjacent lane, while larger values of $\delta$ correspond to safer driving behavior. As expected, $P(S)$ decreases as $\delta$ increases, though this is not linear but rather S-shaped, as shown in \autoref{DrivingComp} for $P(S)$ values at $d =$ 1 km. This figure also reveals that while for $\delta$ = 0.4 s the vehicle is almost certain to reach the right lane before traveling 1 km, as $\delta$ increases the reduction in $P(S)$ is sharper for a higher number of lanes. \par
	
\begin{figure}[t!]
		\centering
		\includegraphics[width = \columnwidth]{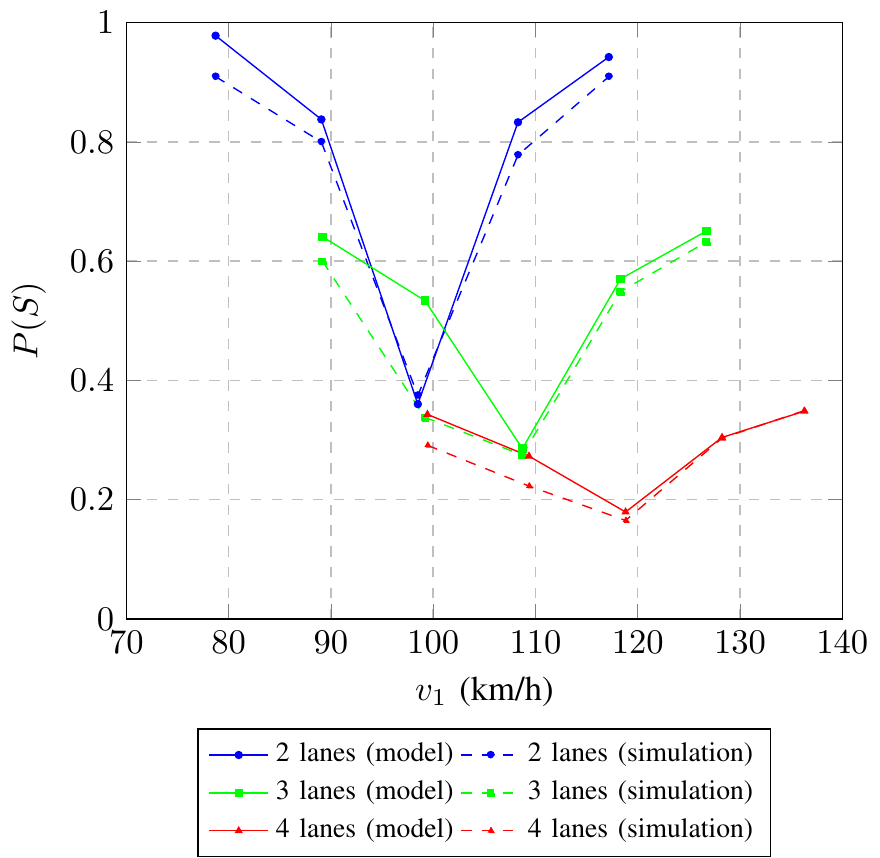}
		\caption{Success probability $P(S)$ as a function of average leftmost lane velocity $v_{1}$ at a distance $d$ of 1 km.} \label{VelocityComp}
	\end{figure}

	\autoref{Velocity} shows $P(S)$ for different values of the leftmost lane average velocity $v_{1}$ (initial ego vehicle velocity) relative to a constant $v_{2}$ (as defined in \autoref{Simulation}) along a 5 km distance for cases with 2, 3, and 4 lanes. Apart from the case where the average velocity of lane 1 is equal to or near that of lane 2, average leftmost lane velocity has a small impact on $P(S)$. This can be seen more clearly in \autoref{VelocityComp} for $P(S)$ values at $d =$ 1 km. As expected, $P(S)$ is lowest when the average velocities of lanes 1 and 2 are equal, i.e. when vehicles in the two lanes do not move relative to each other. As the average velocity of lane 1 increases or decreases relative to lane 2, $P(S)$ increases because of increasing $d_{r}$. However, this rise slows down soon since $d_{r}$ is a function of $v_{1}/v_{2}$.
	
	Finally, an important aspect of the proposed probability model is its near real-time performance. On the computer described in \autoref{Comp}, running the MATLAB\textsuperscript{\textregistered} script to calculate $P(S)$ along a distance $d$ of 5 km took 53, 94, and 143 ms for cases with 2, 3, and 4 lanes, respectively. In general, the probability model is of $\mathcal{O}(nd\ln(d))$. This is important, as it illustrates an efficient implementation and that the model can provide information about near-term goals to the driver or autonomous vehicle at only a fraction of a second (we expect this to be much faster if implemented on the hardware used in autonomous vehicles). More importantly, computation time increases linearly with the number of lanes. This means that even on roads with a high number of lanes, the model can still have near real-time performance.
	
	\section{Conclusions} \label{Conclusion}
	
	This paper presented a model to estimate the probability that a vehicle can reach a near-term goal state using one or multiple lane changing maneuvers. It was shown that for the case of two lanes, the original problem could be simplified to an abstract form with fewer parameters where probability values could be calculated numerically. These values were then used in a recursive method based on the law of total probability to obtain probability values for cases with a higher number of lanes. To validate the probability model and study the effect of different parameters, including distance to the target position, the number of lanes, and traffic density per lane, on the probability, extensive traffic simulations were carried out using VISSIM\textsuperscript{\textregistered}. For most cases, simulation results were within 4\% of model predictions, validating the model. We also discussed the sources of error between model predictions and simulation results and showed that an implementation of the model has near real-time performance. Overall, we conclude that this probability model provides accurate and timely information about reaching a near-term goal state using one or multiple lane changes. \par
	
	Future work will focus on applications of this probability model. We plan on studying the impact of a warning system based on this model on freeway traffic flow during recurrent and non-recurrent congestion using traffic simulations. We also plan on studying the effect of such systems on human driving behavior using a full-cabin driving simulator. Finally, we will pursue applications involving path planning and cooperative navigation for autonomous vehicles.
	
	\section*{Acknowledgment} \label{Section7}
	
	The authors wish to express their gratitude to Dr. Harpreet S. Dhillon for his help with the probability model and to Dr. Montasir Abbas and Awad Abdelhalim for their assistance with VISSIM\textsuperscript{\textregistered} simulations.
	
	\bibliography{BIB}
	\bibliographystyle{IEEEtran}

	\vspace*{-2\baselineskip}
	\begin{IEEEbiography}[{\includegraphics[width = 1 in, height = 1.25 in, clip, keepaspectratio]{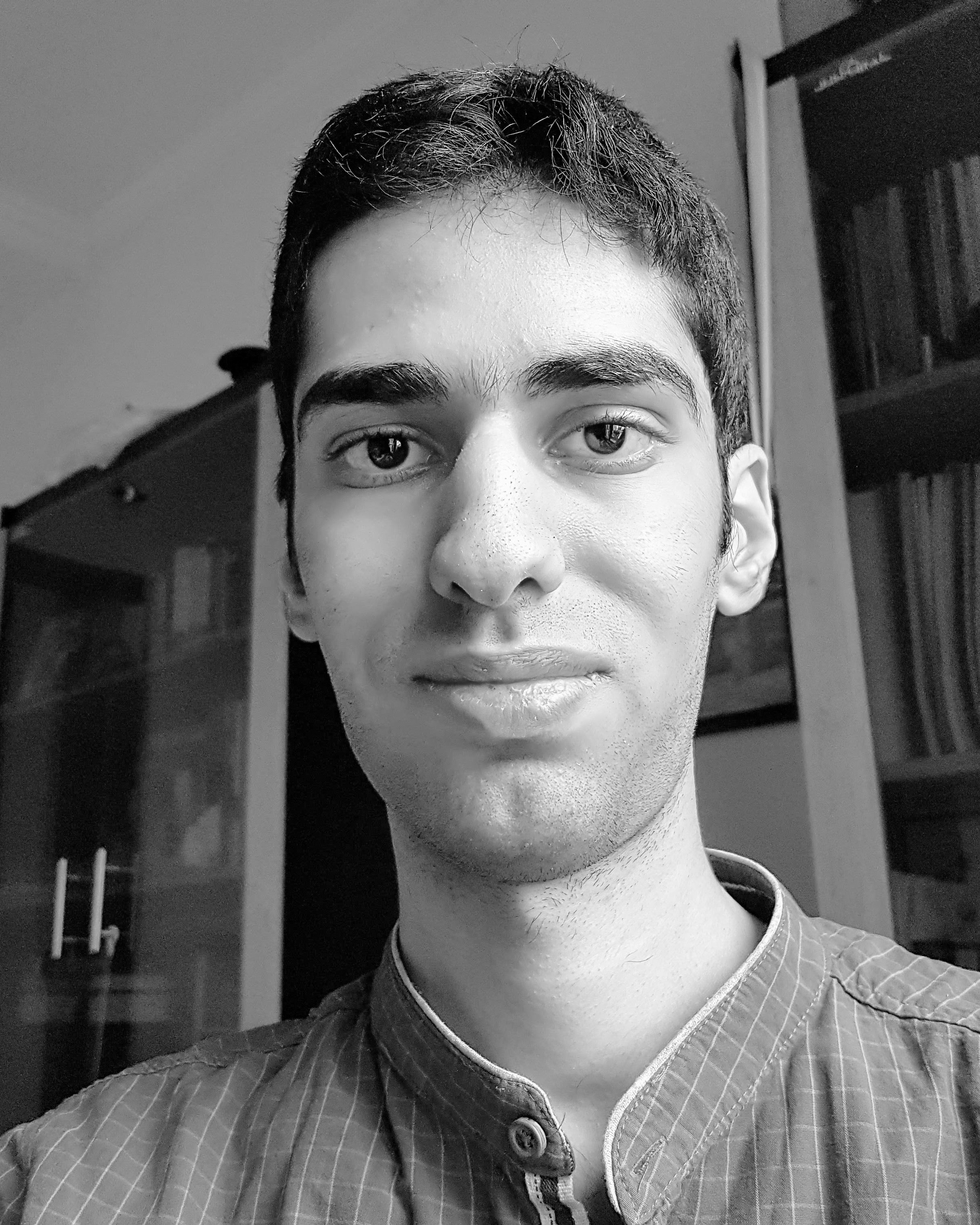}}]{Goodarz Mehr}
	
		received his B.Sc. degree in mechanical engineering from Sharif University of Technology, Tehran, Iran, and is currently pursuing a Ph.D. degree in mechanical engineering from Virginia Tech. His research interests include robotics and control, stochastic planning models, machine learning, and cooperative perception.
		
	\end{IEEEbiography}
	\vspace*{-2\baselineskip}
	\vfill
	\newpage
	\begin{IEEEbiography}[{\includegraphics[width = 1 in, height = 1.25 in, clip, keepaspectratio]{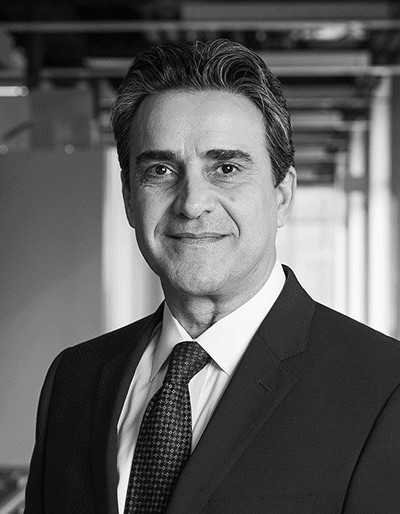}}]{Azim Eskandarian}
	
		received his B.S. degree from George Washington University (GWU), his M.S. degree from Virginia Tech, and his D.Sc. degree from GWU, all in mechanical engineering. He was a Professor of Engineering and Applied Science with GWU and the Founding Director of Center for Intelligent Systems Research from 1996 to 2015, Director of the Transportation Safety and Security University Area of Excellence from 2002 to 2015, Co-Founder of the National Crash Analysis Center in 1992, and Director of the National Crash Analysis Center from 1998 to 2002 and 2013 to 2015. He was an Assistant Professor with Pennsylvania State University, York, PA, USA, from 1989 to 1992, and an Engineer/Project Manager in industry from 1983 to 1989. He has been a Professor and the Head of the Mechanical Engineering Department at Virginia Tech (VT), since 2015. He became the Nicholas and Rebecca Des Champs Chaired Professor in 2018. He established the Autonomous Systems and Intelligent Machines Laboratory at VT to conduct research on intelligent and autonomous vehicles and mobile robots. Dr. Eskandarian is a Fellow of ASME and a member of SAE professional societies. He is a Senior Member of IEEE and received the IEEE ITS Society’s Outstanding Researcher Award in 2017 and GWU’s School of Engineering Outstanding Researcher Award in 2013.
		
	\end{IEEEbiography}
	\vspace*{-2\baselineskip}
	\vfill
	
\end{document}